\documentclass[twocolumn,3p,times]{elsarticle}
\usepackage[utf8]{inputenc}
\usepackage[T1]{fontenc}

\usepackage{graphicx}
\usepackage{svg}
\usepackage{tikz}
\newcommand*\circled[1]{\tikz[baseline=(char.base)]{
            \node[shape=circle,draw,inner sep=2pt] (char) {#1};}}
\usepackage{amsmath,amssymb,amsfonts}%
\usepackage{amsthm}%
\usepackage{algorithm}%
\usepackage{algorithmicx}%
\usepackage{algpseudocode}%
\usepackage{cleveref}
\usepackage{pbox}
\usepackage{makecell}
\usepackage{soul}
\usepackage{xcolor}

\newcommand{\datasetee}{E2E NLG}
\newcommand{\datasetdart}{DART}

\title{Self-training from Self-memory in Data-to-text Generation}

\usepackage{xurl} 

\usepackage{pbox}

\usepackage{amssymb}
\usepackage{svg}
\usepackage{booktabs}  
\usepackage{seqsplit}

\journal{Knowledge-Based Systems}

\begin{document}

\begin{frontmatter}






\author[aff1]{Hoang-Thang Ta~}
\ead{thangth@dlu.edu.vn}



\address[aff1]{Department of Information Technology, Dalat University, Da Lat, Vietnam}

\begin{abstract}

This paper introduces a novel training model, self-training from self-memory (STSM) in data-to-text generation (DTG), allowing the model to self-train on subsets, including self-memory as outputs inferred directly from the trained models and/or the new data. The quality of self-memory is validated by two models, data-to-text (D2T) and text-to-data (T2D), by two pre-defined conditions: (1) the appearance of all source values in the outputs of the D2T model and (2) the ability to convert back to source data in the outputs in the T2D model. We utilize a greedy algorithm to generate shorter D2T outputs if they contain all source values. Subsequently, we use the T2D model to confirm that these outputs can capture input relationships by demonstrating their capacity to convert text back into data. With 30\% of the dataset, we can train the D2T model with a competitive performance compared to full training in the same setup. 
We experiment with our model on two datasets, \datasetee{} and \datasetdart{}. STSM offers the D2T model a  generalization capability from its subset memory while reducing training data volume. Ultimately, we anticipate that this paper will contribute to continual learning solutions that adapt to new training data, incorporating it as a form of self-memory in DTG tasks. The curated dataset is publicly available at: \url{https://github.com/hoangthangta/STSM}.

\end{abstract}





\end{frontmatter}


\section{Introduction}

Within the broad scope of natural language generation (NLG), DTG processes a problem involving converting structured data, typically presented as tables, meaning representations (MRs), knowledge graphs, or sets of triples, into natural language texts. The generated output can be either a summary of the input or a coherent text that adheres to specified criteria, including fluency, naturalness, and incorporating all source values.

A prevalent approach to generating text from a set of triples involves employing a sequence-to-sequence architecture designed for handling sequential data. This often includes using encoder-decoder recurrent neural networks like Transformers. Specifically, it is transfer learning from pre-trained models such as BART and T5. Alternatively, recent methods leverage language instructions and large language models like Alpaca~\cite{taori2023stanford}, ChatGPT~\cite{wu2023brief}, Flan-T5~\cite{longpre2023flan}, and Llama~\cite{touvron2023llama}. While these large models demonstrate relatively strong performance across various DTG tasks, it's important to note that they demand massive memory and supercomputing resources, making them unsuitable for downstream tasks.


In this paper, we introduce a novel model as part of our exploration for an optimization strategy tailored for self-training in D2T, aiming to uphold model performance at a level comparable to full-data training. In each epoch, our model trains on a subset of the training set, incorporating a mix of self-memory derived from trained models and/or newly introduced data. This methodology enables us to reduce the volume of training data in each epoch while maintaining performance levels consistent with full training. Our model also provides helpfulness in continual learning studies, specifically when initiating training on new data from an existing dataset, as opposed to amalgamating new data with the old and training them collectively at the outset. In experiments, we train D2T and T2D models using 30\% of the training set in each epoch, with the optional self-training in the T2D model. Ultimately, we compare the outputs of the D2T model with those of other widely adopted methods to clarify the effectiveness of our model over two datasets, \datasetdart{} and \datasetee{}.

The main contributions of our work are as follows:
\begin{itemize}
    \item \textit{STSM (Self-training from Self-memory)}: The novel training model from the merge of self-memory and/or the new data to keep the output quality on par with the full data training. 
    \item \textit{Data self-training creation}: Provide a step for selecting source-target pairs inferred from the D2T and T2D models by pre-defined criteria. Then, use different combinations of self-memory and/or the new data to create self-training subsets. 
\end{itemize}

The rest of this document is structured as follows: Section 2 discusses previous NLG research, especially DTG problems. Section 3 and Section 4 supply datasets and methods involving our self-training model. Our experiments and discussions based on the results are performed in Section 5. Section 6 addresses the limitations of our work, and finally, we provide conclusions and outline future work in Section 7.

\section{Related Works} 



Rather than handling diverse inputs like NLG, DTG converts structured or semi-structured data into high-quality outputs as human texts. 
DTG is widely applicable due to its capacity to generate meaningful texts, making it valuable in various practical applications, such as weather reports~\cite{reiter2005choosing}, sport news~\cite{kanerva2019template}, and financial comments ~\cite{uehara2020learning}. 
It also encompasses a considerable number of datasets, including those containing tabular data(MLB~\cite{puduppully2019data}, ToTTo~\cite{su2021plan}, ROBOWIRE~\cite{wiseman2017challenges}, WIKIBIO~\cite{lebret2016neural}), triples (DART~\cite{nan2020dart}, WebNLG), meaning representation (E2E NLG~\cite{duvsek2018findings}). These data formats can be interconverted and are typically represented in a linearized text format for model training.

Traditional approaches utilized pipelines featuring multiple modules~\cite{reiter2007architecture}, rule-based generators~\cite{belz2009system}, or statistical machine translation~\cite{belz2009system,mahapatra2016statistical}. However, they are inflexible when adapting to a wide range of datasets and domains compared to neural approaches. Recent DTG studies commonly employ neural networks, with a prevalent adoption of an encoder-decoder architecture~\cite{cho2014learning,sutskever2014sequence} and additional components. ~\citet{ferreira2019neural} applied Gated-Recurrent Units (GRU), Transformers~\cite{vaswani2017attention} for comparing the model performance between neural pipelines and end-to-end approaches. They concluded that the models are better with explicit intermediate steps in the generation process. The process of linearizing inputs results in the loss of structure in the input data. To address this issue, ~\citet{rebuffel2020hierarchical} introduced a hierarchical model incorporating a hierarchical encoder designed for structured data and utilizing hierarchical attention in the decoder. \citet{puduppully2019data} also applied hierarchical attention and entity memory within an encoder-decoder neural network to enhance model performance compared to baseline methods.

Instead of opting for single-stage generation, some works involve a two-stage generation strategy. The first stage usually prepares better input data or additional data that feeds into the text generation in the second stage. To address low-resource generation challenges, \citet{ma2019key} utilized a two-stage model, first generating key facts and subsequently generating texts.  \citet{wang2021sketch} also implemented a two-stage method, employing an autoregressive pointer network for key token selection and a non-autoregressive model for text generation through iterative insertion and deletion operations. In another study, \citet{puduppully2021datamacro} also worked on a two-stage approach, wherein macro plans are derived from the training data and subsequently utilized in the text generation phase. Another type of two-stage generation involves initially producing outputs and subsequently evaluating and selecting the best one based on predetermined criteria. This type is similar to two-stage summarization~\cite{liu2022brio,zhong2020extractive}. For example, \citet{harkous2020have} integrated a semantic fidelity classifier to fine-tune language models to enhance semantic fidelity in a two-stage generation-reranking approach.


Another line of DTG uses neural networks but applies self-training, usually with a few shot learners, to improve the model generalization and deal better with low-resource content. \citet{mehta2021improving} enhanced the model's generalization ability by using a template-based input representation and a self-training method in a few shot settings on pseudo data. \citet{ke2022curriculum} also used self-training as a few-shot learning with pseudo-labeled data generated by the pre-trained model. Furthermore, they proposed Curriculum-Based Self-Training (CBST) to reduce the side effects of low-quality pseudo-labeled data. \citet{deng2023logen} proposed a unified framework for logical knowledge-conditioned text generation in the few-shot setting. Their approach leverages self-training and samples pseudo-logical forms based on content and structure consistency. \citet{nie2019simple} integrated a language understanding module with self-training iterations to achieve strong equivalence between input data and associated text. They then trained a vanilla sequence-to-sequence neural model on refined data to improve content accuracy and mitigate hallucination. On a different work, \citet{kedzie2019good} introduced an architecture-agnostic self-training approach for sampling novel MR/text utterance pairs, effectively running on the expanded dataset, even simple encoder-decoder models with greedy decoding exhibited the capability to generate semantically correct utterances comparable to state-of-the-art outputs. In our study, we implement self-training on 30\% of the training set in each epoch, encompassing both self-memory and the remaining data, instead of using smaller subsets.

Finally, some works employ cycle training to address both D2T and T2D problems simultaneously. \citet{wang2023faithful} utilized cycle training with two inversely related models for D2T and T2D problems, demonstrating comparable performance to fully supervised approaches with limited labeled data. \citet{polat2023improving} applied cycle training by alternately generating text from an input graph and extracting a knowledge graph, emphasizing consistency between the two. They highlighted cycle training's effectiveness in improving performance on evaluation metrics related to syntactic and semantic relations while reducing erroneous output. \citet{guo2020cyclegt} introduced CycleGT, which utilizes fully non-parallel graph and text data to bootstrap and iteratively back-translate between the two forms. Their unsupervised model, trained with the same amount of data, achieves performance on par with several fully supervised models, presenting an effective solution to address data scarcity. While using both D2T and T2D models, we exclusively use the T2D model to validate the output quality (self-memory) of the D2T model, then integrate this memory with new data for self-training the D2T model, with the possibility of optional self-training on the T2D model.


\section{Datasets}\label{sec_dataset}

In this section, we selected two datasets for our experiments, \datasetdart{} and \datasetee{}. Both are medium-sized datasets wherein inputs consist of MRs or tables, subsequently transformed into concise texts. While \datasetdart{} exhibits diversity in its data, \datasetee{} specifically centers around restaurant-related content.

\subsection{\datasetdart{}}

\datasetdart{} (DAta-Record-to-Text) dataset comprises over 82,000 examples aimed at tackling the D2T generation problem, where triplesets serve as inputs and sentences as outputs~\cite{nan2020dart}. Triplesets were extracted from tables, exploiting semantic dependencies among table headers and titles.  The diverse domain dataset is a merge from WikiSQL, WikiTableQuestions, WebNLG, and Cleaned E2E. It is categorized into training (62,659 examples), development (6,980 examples), and testing (12,552 examples). Due to its open-domain nature and ontology-preserving structure, \datasetdart{} poses challenges for state-of-the-art D2T models.

\subsection{\datasetee{}}
E2E NLG is a dataset for producing short texts from restaurant-related meaning representations (MRs)~\cite{novikova2017e2e}. Outputs are lexically rich and syntactically varied, requiring content selection during generation. Models trained on this dataset are expected to yield more natural, varied, and less template-like references. \datasetee{} comprises 50,000+ dialogue-act-based MR combinations, averaging 8.1 references each. Data is split into training, validation, and testing sets (82:9:9 ratio) with similar distributions and distinct MRs. Picture-based data collection captures more natural and informative human references compared to textual MRs~\cite{novikova2016crowd}.

\subsection{Input Linearization}
\label{input_linearization}
For sequence-to-sequence models, it's necessary to have target-source pairs where both elements are in string format. This requires converting inputs into text, a process known as input linearization. \Cref{tab:dataset_examples} displays two examples corresponding to two D2T datasets, \datasetdart{} and \datasetee{} with our input linearization. 
\begin{table*}[!htb]
\small
\centering
\caption{Examples of two D2T datasets with our input linearization.}
\begin{tabular}{lp{10cm}}
\hline
Dataset & Example \\
\hline
DART & \textbf{tripleset:} [ [ ``Clapham'', ''STARTED'', ``20 August'' ], [ ``Clapham'', ``ENDED'', ``20 November'' ], [ ``Clapham'', ``LOAN\_CLUB'', ``Wolverhampton Wanderers'' ] ] \\ \\
 & \textbf{source:} Clapham : STARTED : 20 August \texttt{|} Clapham : ENDED : 20 November \texttt{|} Clapham : LOAN\_CLUB : Wolverhampton Wanderers \\ \\
 & \textbf{target:} \textit{Clapham was loaned by the Wolverhampton Wanderers from 20 August to 20 November} \\
\hline
E2E NLG & \textbf{meaning representation:} name[The Golden Curry], food[English], customer rating[5 out of 5], area[riverside], familyFriendly[yes], near[Café Rouge]\\ \\
 & \textbf{source:} name : The Golden Curry \texttt{|} food : English \texttt{|} customer rating : 5 out of 5 \texttt{|} area : riverside \texttt{|} familyFriendly : yes \texttt{|} near : Café Rouge \\ \\
& \textbf{target:} \textit{The Golden Curry, a 5-star family friendly breakfast joint near the Café Rouge and near the river.} \\
\hline
\end{tabular}
\label{tab:dataset_examples}
\end{table*}

MRs/Triples are concatenated by each other by ``\texttt{ | }'', and elements of each are concatenated by ``\texttt{ : }''. \datasetdart{} offers a group of triples that connect each triple subject to another triple object. Therefore, each triple takes the form of ``\texttt{s : p : o}''. \datasetee{} contains MRs with the common subject. Hence, we remove it in all triples. Each triple remains only two elements \texttt{p} and \texttt{o} in the form of ``\texttt{p : o}''.  We also insert the common subject in the first position of the input string. For example, ``\texttt{name : The Golden Curry}'' refers to the subject ``The Golden Curry'' of the E2ENLG example in \datasetee{}. In this paper, ``\texttt{s}'', ``\texttt{subject name}'', ``\texttt{o}'', and ``\texttt{qv}'' are called \textbf{\textit{source values}} that any generated target must contain.

\section{Methodology}
\subsection{Task Description}


Let $\textbf{x} = {(s_0, p_0, o_0), (s_1, p_1, o_1),...,(s_N, p_N, o_N)}$ represent a set of triples with a size of $N$. In each triple $(s,p,o)$, $s$ denotes a subject, $o$ denotes an object, and $p$ signifies the relationship between $s$ and $o$. If $\textbf{x}$ is a set of MRs, we express it as $\textbf{x} = {(k_0, v_0), (k_1, v_1),...,(k_N, v_N)}$ with a size of $N$, where each element is a key-value pair. Then, let $\textbf{y}=\{w_0, w_1,...,w_M\}$ be a representation text for $\textbf{x}$ with $M$ symbols.

Next, $\textbf{x}$ is transformed into a string in the linearization process as in \Cref{input_linearization}. There are two tasks: D2T and T2D. The first one involves converting an input $\textbf{x}$ to $\textbf{y}$. Conversely, in the T2D task, when provided with an input $\textbf{y}$, the model is expected to convert it to $\textbf{x}$.

\subsection{Transformers}
We apply Transformers~\cite{vaswani2017attention}, a sequence-to-sequence architecture, to set up D2T and T2D models. A Transformer contains an encoder, a decoder, and an attention mechanism. Given a source $X=\{x_1, x_2, x_3, ..., x_N\}$ with $N$ symbols and a target $Y=\{y_1, y_2, y_3,...,y_M\}$ with $M$ symbols. The encoder yields a representation $Z = \{z_1, z_2, z_3,..., z_N\}$ from $X$ with the same number of symbols. Later, the decoder takes $Z$ to produce the target $Y$. The chain rule probability $p(Y|Z)$ to generate $Y$ from $Z$ is:
\begin{equation}
\begin{aligned}
p(Y|Z) &= \prod_{i}^{M}p(y_i|Y_{<i},Z)
\end{aligned}
\label{eq:chain_prob}
\end{equation}
which $y_0$ is the ``start'' symbol (\texttt{<bos>}) and $Y_{<i}$ is a sequence of previous symbols of $y_i$. When meeting the      ``end'' token (\texttt{<eos>}) or the maximum length, the inference process ends. The cross-entropy loss $L_{ent}$ minimizes the sum of negative loglikelihoods of the symbols:
\begin{equation}
\begin{aligned}
L_{ent} = - \sum^{M}_{j=1}\sum^{}_{w}p_{true}(w|Y_{<j},Z)log( p(w|Y_{<j},Z))
\end{aligned}
\label{eq:entropy_loss}
\end{equation}
which $p_{true}$ is a one-hot distribution:

\begin{equation}
\begin{aligned}
p_{true}(w|Y_{<j},Z) = \bigg\{
\begin{matrix}
1 \;\;\;\;\; w = y_j 
\\
0 \;\;\;\;\; w \neq y_j
\end{matrix}
\end{aligned}
\label{eq:one_hot_dis}
\end{equation}


A Transformer has two attention functions: Scaled Dot-Product Attention and Multi-Head Attention. Let $Q$, $K$, $Q$ be the query matrix, the key matrix, and the value matrix correspondingly. Let $d_k$, $d_k$ be the dimensions of queries and keys, and $d_v$ be the dimension of values. The attention function of Scaled Dot-Product Attention is ~\cite{vaswani2017attention}:
\begin{equation}
\begin{aligned}
Attention(Q,K,V) = softmax(\frac{QK^T}{\sqrt{d_k}})V
\end{aligned}
\label{eq:attention}
\end{equation}
%

The purpose of using the scaling factor $\frac{1}{\sqrt{d_{k}}}$ is to avoid the softmax function from experiencing very small gradients when the value of $d_k$ becomes substantial. Additionally, the Multi-Head Attention mechanism operates with keys, values, and queries, each having a dimension of $d_{model}$~\cite{vaswani2017attention}. This setup enables the model to learn additional information from different positions' subspace representations.

\begin{equation}
\begin{aligned}
MultiHead(Q,K,V) &= Concat(head_1,head_2,...,head_h)W^O \\
\text{where  } head_i &= Attention(QW_i^Q,KW_i^K,VW_i^V)
\end{aligned}
\label{eq:multihead_att}
\end{equation}
which $h$ refers to the number of heads. For each head $i$, $W^Q_{i} \in \mathbb{R}^{d_{model} \times d_k}$, $W^K_{i} \in \mathbb{R}^{d_{model} \times d_k}$, $W^V_{i} \in \mathbb{R}^{d_{model} \times d_v}$, $W^O \in \mathbb{R}^{d_{model} \times hd_v}$ are the parameter matrices.

\subsection{Self-training Model}

\begin{figure*}[htbp]
  \centering

\includegraphics[width=1.75\columnwidth]{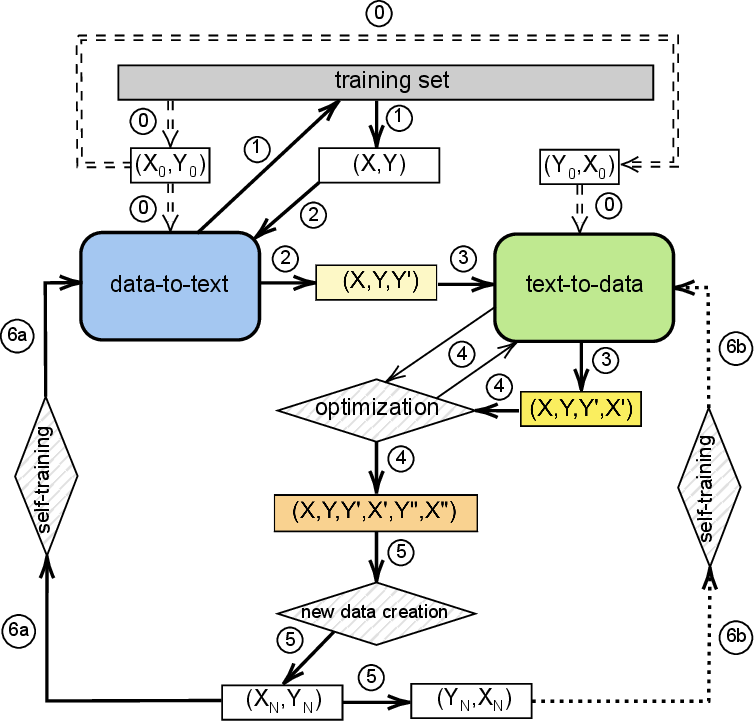}
  \centering
  \caption{The self-training model. \protect\circled{0} gets a fixed/random subset for first training D2T and T2D models; \protect\circled{1} gets a fixed/random subset for starting to self-train D2T and T2D models; \protect\circled{2} \& \protect\circled{3} infer $Y'$ and $X'$; \protect\circled{4} optimizes $Y'$ as $Y''$ and use it to infer $X''$;  \protect\circled{5} creates new data, ($X_N, Y_N$) and ($Y_N, X_N$); \protect\circled{6a} \& \protect\circled{6b} self-train D2T and T2D models on new data. The T2D self-training is optional.}
\label{self_training_model}
\end{figure*}



The self-training framework, described in \Cref{self_training_model}, consists of two sequence-to-sequence models named T2D and D2T. These models engage in self-training using a freshly generated subset, identified based on predetermined criteria and data combinations. This subset encompasses both newly acquired data and specific self-memory chosen during the optimization process.

Let define $D$ is the training set, and $(X,Y)$ is a subset of $D$, including source-target pairs $(x_i,y_i)$. Let's define $(X', Y') $ as a set of target-source pairs $(x'_i,y'_i)$ inferred from the D2T and T2D models. Similarly, $(X'', Y'')$ is a set of source-target pairs $(x''_i,y''_i)$ after passing the optimization step. The self-training model contains these steps:

\begin{itemize}
    \item \circled{0}: First, a subset $(X_0,Y_0)$ is extracted fixedly or randomly from the training set for training the D2T model and its swap version $(Y_0, X_0)$ for the T2D model. This aims to prepare two trained models for inferring self-memory, which is later used for self-training.
    \item \circled{1}: Extract another fixed/random subset $(X, Y)$ from the dataset, and it must has the same size as $(X_0,Y_0)$.
    \item \circled{2} \& \circled{3}: Feed $X$ to the D2T model to get self-memory as a set of output texts $Y'$. Next, feed $Y'$ to the T2D model to get self-memory as a set of output texts $X'$. 
    \item \circled{4}: Pass $Y'$ to the optimization step to get $Y''$, then continue to pass $Y''$ to the T2D model to infer $X''$. At this step, the tuple $(X,Y,Y',X',Y'',X'')$  is obtained.
     \item \circled{5}: From the data in step \circled{4}, we filter \textit{\textbf{``elite''}} pairs that satisfy the pre-defined conditions in \Cref{new_data_creation}. Then, use different strategies to mix these pairs with the original data $(X, Y)$ to prepare two subsets for the next step. For example, we combine $(X,Y)$ with $(X,Y')$ and $(X,Y'')$. Note that the size of subsets always equals the size of $(X,Y)$ in step \circled{1}. If the size exceeds the requirement, we randomly select the number of examples to fit that size.
     \item \circled{6a} \&  \circled{6b}: Use subsets in step \circled{5} for self-training D2T and T2D models. Self-training the T2D model or \circled{6b} is not mandatory.
     \item Repeat from step \circled{1} to \circled{6a} and/or \circled{6b} by the given number of self-training epochs. Finally, we evaluate the output quality of the D2T model.
\end{itemize}

The best D2T and T2D models are saved during each epoch based on metric values evaluated on the validation set. Specifically, the D2T model uses METEOR, while the T2D model uses OSF-precision (Overall Slot Filling)~\cite{wang2018describing}, as presented in detail in \Cref{automatic_metrics}.

\subsection{Target Optimization}


In DTG, an ideal target must capture all information in the input, usually source values and relationships between subjects and objects (keys and values), and have no redundant information or hallucinations. With that sense, we believe that any generated target that satisfies these conditions and is shorter than the gold target should be chosen as the new gold target, or called the optimized target. 


\Cref{tab:optimization_example} displays two examples on \datasetdart{} and \datasetee{}, each contains the optimized target and the gold target. The optimized target is more brief than the gold target and features a different word order. Consequently, we argue that BLUE may not be a suitable metric for evaluating the model's performance. Therefore, we opt for METEOR. Note that the optimized target must validate its ability to capture the relationships in the source before being selected as self-memory for self-training, as shown in \Cref{role_T2D_model}.

\begin{table*}[ht]
\small
\centering
\caption{Two examples, each with the source, the gold target and the optimized target in \datasetdart{} and \datasetee{}. While the gold target and the optimized target capture all source values, the optimized target is shorter in length. The figure is shown better with colors.}. 
\begin{tabular}{p{12cm}}
\hline
\textbf{Example 1 (\datasetdart{})}
\\
\hline
\textbf{Source:} \colorbox{cyan}{Antwerp International Airport} : OPERATING\_ORGANISATION : \colorbox{lime}{Flemish Government}  \texttt{|} \colorbox{cyan}{Antwerp International Airport} : OWNER : \colorbox{orange}{Flemish Region}  \\ \\ 
\textbf{Gold target:} The owner of \colorbox{cyan}{Antwerp International Airport} is the \colorbox{orange}{Flemish Region} and the operating organisation is the \colorbox{lime}{Flemish government}. \\ \\
\textbf{Optimized Target:} \colorbox{cyan}{Antwerp International Airport} is operated by the \colorbox{lime}{Flemish Government} and is owned by the \colorbox{orange}{Flemish Region}. \\ 
\hline
\hline
\textbf{Example 2 (\datasetee{}{})}
\\
\hline
\textbf{Source:} name : \colorbox{lime}{The Mill} \texttt{|} eatType : \colorbox{orange}{coffee shop} \texttt{|} food : \colorbox{cyan}{Indian} \texttt{|} priceRange : \colorbox{pink}{cheap} \texttt{|}  area : \colorbox{teal}{riverside} \texttt{|} near : \colorbox{yellow}{The Sorrento}  \\ \\
\textbf{Gold target:} Near \colorbox{yellow}{The Sorrento}, by the \colorbox{teal}{riverside} there is a \colorbox{orange}{coffee shop} called \colorbox{lime}{The Mill}. It is \colorbox{pink}{cheap} and serves \colorbox{cyan}{Indian} food. \\ \\
\textbf{Optimized Target:} \colorbox{lime}{The Mill} is a \colorbox{pink}{cheap} \colorbox{orange}{coffee shop} that serves \colorbox{cyan}{Indian} food in the \colorbox{teal}{riverside} area near \colorbox{yellow}{The Sorrento}. \\
\hline
\end{tabular}
\label{tab:optimization_example}
\end{table*}

\begin{algorithm}
\caption{Target optimization}\label{algo1}
\hspace*{\algorithmicindent} \textbf{Input}: $(x', y')$ \\
\hspace*{\algorithmicindent} \textbf{Output}: $y''$ 
\begin{algorithmic}[1]
\State $MS, MV \Leftarrow [], []$
\State $S \Leftarrow split\_sentence(y')$
\State $V \Leftarrow extract\_value(x')$
\For{$s$ in $S$}
    \For{$v$ in $V$}
        \If{$v$ in $s$ \& $v$ not in $MV$}
            \State $MV \Leftarrow add(v)$
            \If{$s$ not in $MS$}
                \State $MS \Leftarrow add(s)$
            \EndIf
        \EndIf
    \EndFor
\EndFor
\If {$|$MV$|$ = $|$V$|$}   
    \State $y'' \Leftarrow flatten\_string(MS)$
\Else
    \State $y'' \Leftarrow y'$
\EndIf
\end{algorithmic}
\end{algorithm}

We apply Algorithm 1, a simple greedy algorithm, from a generated target to loop its sentence list to choose sentences containing source values. The input is a pair of source-target $(x', y')$ inferred from the D2T and T2D models, and the output is an optimized target $y'$. The optimized target is expected to capture all source values, each happening only once.

Here is the explanation of how Algorithm 1 works. From $(x', y')$, the algorithm starts with creating two empty lists, the matched sentences $MS$ and the matched values $MV$. It also extracts a list of sentences $S$ from $y'$ and a list of source values $V$ from $x'$. Next, the algorithm uses two loops, one loop for each sentence $s$ in $S$ and one sub-loop for each value $v$ in $V$. In each step, if $v$ appears in $s$ and does not exist in $MV$, add it to $MV$. If $s$ is not in $MS$, add $s$ to $MS$. The algorithm then compares the element numbers of $MV$ and $V$. If they are equal, $MS$ is flattened into a string by concatenating each element with a blank, resulting in a new target $y''$. Otherwise, we have no new target. 

\subsection{New Data Creation}\label{new_data_creation}
From tuple data $(X, Y, Y', X', Y'', X'')$, we filter ``\textit{\textbf{elite}}'' pairs of source-target from each tuple $(x, y, y', x', y'', x'')$ by two cases.

\textbf{Case 1}: If $y''$ is $y'$, there is no new optimized target. A pair $(x, y')$ is chosen if it satisfies these conditions:
\begin{itemize}
    \item \textbf{\textit{1a}}. The length of $y'$ is less than the length of $y$.
    \item \textbf{\textit{1b}}. All source values of $x$ must appear in $y'$. 
    \item \textbf{\textit{1c}}. All MRs/triples of $x'$ must be a subset of those of $x$.  In this case, we apply OSF-precision, as presented in \Cref{automatic_metrics}.
\end{itemize}

\textbf{Case 2}: If $y''$ is not $y'$, it means there has a new optimized target. In this case, we will select a pair $(x, y'')$ if guarantee these conditions:
\begin{itemize}
    \item \textbf{\textit{2a}}. The length of $y''$ is less than the length of $y$.
    \item \textbf{\textit{2b}}. All source values of $x$ must appear in $y''$. 
    \item \textbf{\textit{2c}}. All MRs/triples of $x'$ must be a subset of those of $x$.  In this case, we apply OSF-precision, as presented more \Cref{automatic_metrics}.
\end{itemize}

Let $(X, Y)$ be a set of distinctive pairs $(x, y)$, which is used to feed to the D2T and T2D models to get a tuple $(Y', X,' Y'', X'')$ in \Cref{self_training_model}. Let $(X_O, Y_O)$ represent a set of distinctive pairs $(x_o, y_o)$ obtained by selecting either $y'$ or $y''$ from pairs $(x, y')$ and $(x, y'')$ based on Case 1 or Case 2 decisions. Similarly, let $(X_R, Y_R)$ be the remaining new data with pairs $(x, y)$ where both $y'$ and $y''$ cannot be inferred from $x$. We combine $(X_O, Y_O)$ and $(X_R, Y_R)$ to form a new set $(X_N, Y_N)$. If the number of pairs in $(X_N, Y_N)$ exceeds that of $(X, Y)$, we \textbf{\textit{randomly select pairs}} to match the number of pairs in $(X, Y)$. This ensures an equivalent number of pairs in the subset for self-training. Finally, we have a new subset $(X_N, Y_N)$ with distinctive pairs and the same size as $(X, Y)$.

\subsection{Roles of the T2D model}\label{role_T2D_model}

In DTG, popular methods usually apply a single D2T model to produce a target $y'$ (greedy decoding) or a set of $y'$ (beam search) from a given source $x$. Later, these targets are validated by automatic metrics against the source, such as cosine similarity, ROUGE, and any metric combinations, or based on several pre-defined conditions to produce the quality scores. Though this method helps choose the best target, it remains unknown how a target can capture the relationships between MRs/triples in the source. 

A T2D model allows one to check the relationships the D2T target $y'$ can capture by converting it back to $x'$. An ideal target $y'$ is when its T2D output $x'$ matches $x$ totally regarding MRs/triples. Applying this strict criterion to filter source-target pairs appears to result in a limited amount of new data available for self-training. This could be attributed to the varied and latent relationships the T2D model needs to handle when extracting text information. Instead, we use a looser condition (match a subset of MRs/triples), as presented  \textbf{\textit{1c}} and \textbf{\textit{2c}} in \Cref{new_data_creation}. 
The appearance of all source values of $x'$ in $y$ can somewhat prove the capture of relationships of the target $y'$. In short, the T2D model can validate relationships of MRs/triples in a given D2T target.

\section{Experiments and Results}

\subsection{Automatic Evaluation Metrics}
\label{automatic_metrics}

We use widely used metrics to assess the quality of the produced texts, contrasting them with the references. These metrics can be classified into two essential categories: string-based and semantic-based. In this paper, we only use several string-based metrics such as BLEU, CIDEr, EPM, NIST, OSF, ROUGE, and TER. 


\noindent\textbf{BLEU (Bilingual Evaluation Understudy)}: BLEU provides a resultant score for assessing the resemblance between a produced target and a reference target. It accomplishes this by tallying the shared n-grams while also considering a penalty for brevity based on text length. BLEU is commonly applied in tasks like machine translation, especially when the length of the target text is roughly equivalent to that of the source text~\cite{papineni2002bleu}.

\noindent\textbf{CIDEr (Consensus-based Image Description Evaluation):} This metric evaluates the quality of captions created for images by contrasting them against reference captions supplied by human annotators. CIDEr examines the precision of the generated captions (how closely they align with the reference captions) and their distinctiveness (how dissimilar they are from one another). CIDEr delivers a more thorough evaluation than BLEU and ROUGE~\cite{vedantam2015cider}.

\noindent\textbf{EPM (Exact Phrase Matching):} EMP is a simple metric to determine the quality of the generated target. It computes a score from the number of source values that appear precisely in the generated target over the number of source values extracted from the source. The phrase matching is not always correct when data types can be boolean or ordinal. However, the higher metric score generally reflects a more favorite quality.

\noindent\textbf{METEOR (Metric for Evaluation of Translation with Explicit ORdering):} It computes an F-score between the generated target and the gold target by the overlapped unigrams from three modules based on words: exact, stem, and synonymy~\cite{banerjee2005meteor}. METEOR is superior to BLEU when dealing with low-source texts and obtains a higher human judgment at sentence level~\cite{lavie2009meteor}.

\noindent\textbf{NIST (N-gram-based Integrated Evaluation Score):} This metric is widespread in machine translation for evaluating the caliber of machine-produced translations. NIST calculates a similarity score by comparing the generated translation with the reference translation, relying on the overlap of n-grams. What sets NIST apart from specific alternative metrics is its integration of normalized weights assigned to varying n-gram lengths. This unique feature enables NIST to be flexible and adaptable across diverse languages and translation nuances~\cite{doddington2002automatic}.

\noindent\textbf{OSF (Overall Slot Filling):} It evaluates precision (OSF-precision), recall (OSF-recall), and F-score (OSF-F) by comparing two Knowledge Base (KB) representations. A slot is deemed correct if both the reconstructed KB and the input KB contain a matching pair of slot type and its corresponding slot value.~\cite{wang2018describing}. Our work computes OSF-precision, OSF-recall, and OSF-F1 between D2T inputs and T2D outputs. An MR/triple of a D2T input must match totally that of a T2D output to be considered a match slot.



\noindent\textbf{ROUGE (Recall-Oriented Understudy for Gisting Evaluation):} ROUGE is a traditional metric for summarization problems, computing the degree of similarity between a generated target and a gold target relying on the overlap rate of the number of semantic units (n-grams) appearing in both texts~\cite{lin2004rouge}. Several popular ROUGE types include ROUGE-1, ROUGE-2, and ROUGE-LCS (Longest Common Subsequent).

\noindent\textbf{TER (Translation Edit Rate):} It  measures machine translation quality by quantifying edits needed to align the machine-generated translation with a human reference. Effective for assessing translations with substantial differences in word order, grammar, and structure, TER focuses on structural alignment. Lower TER scores indicate greater similarity between machine-generated and reference translations. However, TER doesn't account for nuances like meaning, word choice, or fluency, emphasizing structural alignment~\cite{snover2009fluency}.

\subsection{Experimental configurations}
\label{experiment_config}

\begin{figure*}[htbp]
  \centering
\includegraphics[width=1.75\columnwidth]{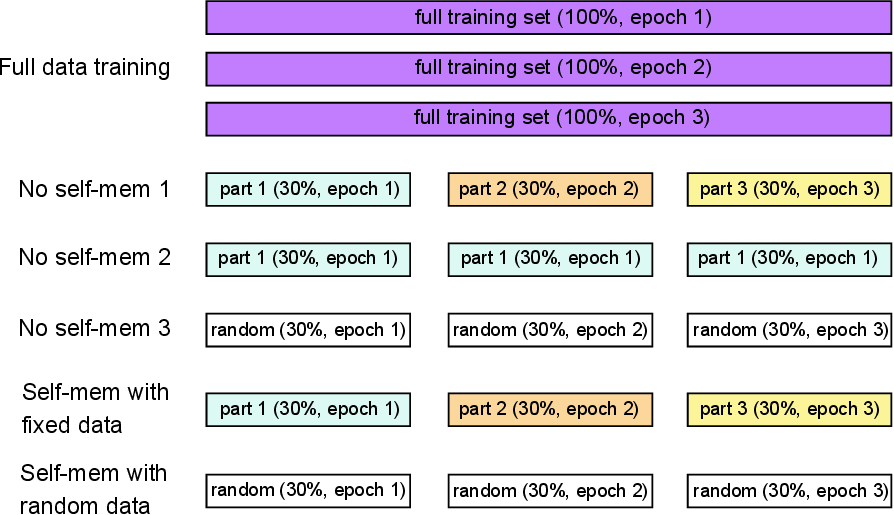}
  \centering
  \caption{The data allocation for different training methods. The figure is shown better with colors.}
\label{data_training_methods}
\end{figure*}

In all methods, the training process involves 3 epochs, with a maximum input/output length of 256, a minimum input/output length of 4, an adaptive learning rate, and greedy decoding in the inference time. Transfer learning is applied to train Transformer models on the full training data, leveraging pre-trained models like BART-base, FLAN-T5-base, and T5-base. T5-base is utilized in all self-training methods to train on both random and fixed data.
\begin{itemize}
    \item \textit{Fixed data}: The training set is divided into non-overlapping subsets, each employed in every training epoch. In experiments, we divide the training set into three parts, each comprising 30\% data. It is possible to increase or decrease this data percentage, but when 30\% of training data helps the model earn a competitive performance. In total, there is 90\% data for 3 epochs, and leave the remaining 10\% unused. 
    \item \textit{Random data}: Each epoch takes 30\% random training data.
\end{itemize}

In this situation, it is evident that self-training methods utilize less data than full training with equivalent epochs. To be precise, a self-training method necessitates only 90\% training data for 3 epochs, whereas a full-training method demands 300\% training data for the same 3 epochs (with each utilizing 100\% training data or the whole training set). Besides the type of training data, we also set up several different training methods:
\begin{itemize}
    \item \textit{no self-mem}: The training has no self-memory. It is similar to training the models over the training set but takes its subsets. There are three types of no-self-memory training. The first type (\textit{no self-mem 1}) takes non-overlap fixed subsets in each epoch. The second type (\textit{no self-mem 2}) takes the same subset for all epochs. The third type takes a random subset for each epoch. The first and second types are used for fixed data training, while the third one (\textit{no self-mem 3}) is for random data training.
    \item \textit{self-mem}: The training contains two steps. The first step is to train data for creating the T2D and D2T models. Then, in the second step, use these models to infer self-memory for self-training the D2T model.
    \item \textit{self-mem + self T2D}: Similar to \textit{self-mem} but this setup allows self-training of the T2D model as well.
    \item \textit{self-mem + new data}: Similar to \textit{self-mem}, the training takes self-memory and new data for self-training the D2T model but not the T2D model.
    \item \textit{self-mem + new data + self T2D}: Similar to \textit{self-mem}, the training takes self-memory and new data for self-training the D2T and T2D models.
\end{itemize}


\Cref{data_training_methods} illustrates the data allocation across 3 epochs for various training methods. The full training method requires the largest dataset, encompassing the entire training set (100\% data) for each epoch. Both \textit{no self-mem 1} and \textit{self-mem with fixed data} employ identical training data divided into three segments for each training epoch. Similarly, \textit{no self-mem 3} and \textit{self-mem with random data} utilize random subset data for each training session. \textit{no self-mem 2} exclusively utilizes the first portion of the training set for training across 3 epochs, so this method has less data diversity than others.

\subsection{\datasetdart{}'s results}

\begin{table*}[!htb]
\small
\centering
\caption{Metric values between the generated targets and the gold targets on the test set of \datasetdart{} by different methods.}

\begin{tabular}{llll}
\hline
\textbf{Methods} & \textbf{BLEU}$\uparrow$ & \textbf{METEOR}$\uparrow$ & \textbf{TER}$\downarrow$ \\
\hline
LSTM with Attention (\citet{nan2020dart}) &   29.66 &  27 &  63 \\
End-to-End Transformer (\citet{nan2020dart}) &   27.24 &  25 &  65 \\
BART-base (\citet{nan2020dart}) &  47.11 &  38 &  46 \\
T5-base (\citet{nan2020dart}) &  \textbf{49.21} & \textbf{40} &  \textbf{44} \\
\hline
\multicolumn{4}{l}{{Our run}} \\
BART-base & 46.15 & 38.13  & 48.20 \\
FLAN-T5-base & 45.53 & 0.36 & 0.54  \\
T5-base & 48.52 & 39.70 & 45.91  \\
\hline
\multicolumn{4}{l}{30\% fixed data per epoch$^{1}$} \\
\textit{no self-mem 1} & 44.86 & 37.98 & 50.83  \\
\textit{no self-mem 2}& 41.23 & 33.98 & 51.47 \\

\textit{self-mem} & 45.63 & 38.90 & 49.10 \\
\textit{self-mem + self T2D} & 46.14 & 38.86 & 49.08  \\ 
\textit{self-mem + new data} & 47.76 & 39.51 & 48.28 \\
\textit{self-mem + new data + self T2D} & 47.60 & 39.39 & 48.23 \\

\hline
\multicolumn{4}{l}{30\% random data per epoch} \\
\textit{no self-mem 3} & 46.90 & 38.92 & 46.61  \\
\textit{self-mem} & 44.52 & 38.62 & 50.75\\
\textit{self-mem + self T2D} & 45.62 & 38.81 & 48.95  \\
\textit{self-mem + new data} & 47.54 & 39.38 & 48.07  \\
\textit{self-mem + new data + self T2D} & 47.32 & 39.19 & 47.34  \\
\hline
\multicolumn{4}{l}{\makecell[l]{$\uparrow$: higher is better, $\downarrow$: lower is better \\
$^{1}$A non-overlap data is allocated in each epoch, except \textit{no self-mem 2}}} \\
\hline
\end{tabular}
\label{tab:dart_results}
\end{table*}

To ensure an equitable comparison, we employed the evaluation package from the challenge\footnote{https://github.com/Yale-LILY/dart/tree/master/evaluation}, to assess BLEU, METEOR, and TER values on the test set. Our methods are compared to:
\begin{itemize}
    \item LSTM with Attention: This model is a Bidirectional Long Short-Term Memory (Bi-LSTM) model featuring an attention mechanism~\cite{nan2020dart}. The encoder comprises a 2-layer Bi-LSTM with 300-dimensional word embeddings and does not rely on pre-trained word vectors. For the decoder, the configuration includes 512 hidden units and a dropout rate of 0.3.
    \item End-to-End Transformer: \citet{nan2020dart} utilized a Transformer architecture~\cite{vaswani2017attention} for training their models, and we adopted their results.
    \item BART-base: Our models underwent training on BART-base, a version of the BART model that incorporates a denoising autoencoder for the pretraining of sequence-to-sequence models \cite{lewis2019bart}. Additionally, we utilized the results obtained from BART-base in the study by \citet{nan2020dart}.
    \item T5-base: In a similar way, our models were trained on T5-base \cite{raffel2020exploring}, and we took the results obtained by \citet{nan2020dart} for the purpose of comparison.
\end{itemize}

\Cref{tab:dart_results} shows the results of the test set using various methods evaluated through widely used string metrics such as BLEU, METEOR, and TER. T5-base, as reported by \citet{nan2020dart}, demonstrated the highest performance, whereas our version exhibited a slightly lower performance but outperformed our FLAN-T5-base and BART-base models. In no self-memory training methods, \textit{no self-mem 3} trained on random data outperformed the other two methods (\textit{no self-mem 1} and \textit{no self-mem 2}), which were trained on fixed data. Among the self-memory methods, training on fixed data generally yielded slightly better results than training on random data. Specifically, \textit{self-mem + new data} on fixed data emerged as the optimized method, achieving highly competitive results compared to the best method over T5-base. Additionally, it was observed that self-training on the T2D model did not significantly contribute to performance improvement. Therefore, we recommend excluding this self-training to conserve time and computer resources.

\subsection{\datasetee{}'s results}
\begin{table*}[!htb]
\small
\centering
\caption{Metric values between the generated and targets on the test set of \datasetee{} by different methods.}

\resizebox{\textwidth}{!}{
\begin{tabular}{lllllll}
\hline
\textbf{Methods} & \textbf{BLEU}$\uparrow$ & \textbf{ME}$\uparrow$ & \textbf{NIST}$\uparrow$ & \textbf{ROUGE-L}$\uparrow$ & \textbf{CIDEr}$\uparrow$  \\
\hline
Pragmatics~(\citet{shen2019pragmatically}) & \textbf{68.60} & 45.25 & \textbf{8.73} & \textbf{70.82} & \textbf{2.37} \\
EDA\_CS~(\citet{roberti2020copy}) & 67.05	 & 44.49 & 8.51 & 68.94 & 2.23 \\
SLUG~(\citet{juraska2018deep}) & 66.19 & 44.54 & 8.61 & 67.72 \\
TGEN~(\citet{duvsek2018findings}) & 65.93 & 44.83	& 8.60 &	68.50 &	2.23 \\
\hline
\multicolumn{5}{l}{Our run} \\
BART-base & 65.74 & 45.60 & 8.46 & 68.76 & 2.20 \\
FLAN-T5-base & 65.65 & 45.54 & 8.49 & 67.85 & 2.12 \\
T5-base & 66.95 & 45.70 & 8.59 & 68.97 & 2.27 \\
\hline
\multicolumn{5}{l}{30\% fixed data per epoch$^{1}$} \\
\textit{no self-mem 1} & 65.47 & 45.84 & 8.32 & 68.33 & 2.17 \\
\textit{no self-mem 2} & 64.94 & 45.13 & 8.33 & 67.76 & 2.21 \\
\textit{self-mem}  & 61.28 & 44.84 & 8.05 & 66.58 & 2.05 \\
\textit{self-mem + self T2D}  & 60.69 & 44.48 & 8.03 & 66.45 & 2.03 \\ 
\textit{self-mem + new data} & 65.55 & 46.07 & 8.35 & 68.16 & 2.10\\
\textit{self-mem + new data + self T2D} & 
65.47 & 46.10 & 8.38 & 68.11 & 2.07 \\
\hline
\multicolumn{5}{l}{30\% random data per epoch} \\
\textit{no self-mem 3} &  65.50 & 45.30 & 8.40 & 68.24 & 2.20 \\
\textit{self-mem} & 61.98 & 44.48 & 8.05 & 66.44 & 2.14 \\
\textit{self-mem + self T2D} & 61.74 & 44.89 & 8.07 & 66.54 & 2.11
\\
\textit{self-mem + new data} & 65.11 & \textbf{46.11} & 8.35 & 68.41 & 2.08 \\
\textit{self-mem + new data + self T2D} & 65.55 & 45.69 & 8.41 & 68.45   & 2.16 \\
\hline
\multicolumn{6}{l}{\makecell[l]{$\uparrow$: higher is better \\ $^{1}$A non-overlap data is allocated in each epoch, except for \textit{no self-mem 2}}} \\
\hline
\end{tabular}
}
\label{tab:e2e_results}
\end{table*}

Similar to the evaluation on \datasetdart{}, we use the original package from E2E NLG Challenge\footnote{https://github.com/tuetschek/e2e-metrics} to have a fair comparison between our methods and others. This package measures the output quality by automatic metrics such as BLEU, NIST, METEOR, ROUGE-L, and CIDEr. We compare our self-memory methods to the training of full models and other benchmark methods, including:
\begin{itemize}
    \item Pragmatics~\cite{shen2019pragmatically}: Two practical approaches to text generation focusing on retaining information and explicitly distractor elements.
    \item EDA\_CS~\cite{roberti2020copy}: A sequence-to-sequence neural network with attention and copy mechanisms at a character level.
    \item SLUG~\cite{juraska2018deep}: A collaborative neural language generation incorporating novel techniques for enhancing data representation and augmentation.
    \item TGEN~\cite{duvsek2018findings}: A sequence-to-sequence model with an attention mechanism, decoding beam search, and a reranking mechanism on generated outputs.
\end{itemize}

Similar to the training on \datasetdart{}, T5-base outperformed the other two, FLAN-T5-base and BART-base. Furthermore, the self-memory training shows worse when lacking both new data and self-training of T2D.  Self-training on new data helps to improve the output quality, while self-training on the T2D model does not contribute significantly. The best METEOR score of 46.11 is achieved by \textit{self-mem + new data} when incorporating newly added random data. On the other hand, its model trained on fixed data exhibits a competitive METEOR score of 46.07 and a competitive BLUE. The best model is Pragmatics~\cite{shen2019pragmatically} with a BLUE score of 68.60. Besides, the no self-memory models on subsets obtained a competitive performance compared to full-training and self-memory models.

\subsection{General results}

With results over \datasetdart{} and \datasetee{}, we realize that \textit{self-mem + new data} is the optimized self-memory method which has a competitive performance compared to the full data training. It combines self-memory and new data, with self-training the D2T model and without self-training the T2D model. Moreover, self-training on the T2D model is unnecessary, and we only need to train the T2D model once to check the quality of texts inferred from the D2T model. The superiority between self-training on fixed data and random data remains uncertain. Nonetheless, opting for fixed data in training is more advantageous, particularly in scenarios involving continual learning with the introduction of new data.

\section{Limitation}
The first limitation pertains to our model's evaluation, which was confined to small-scale pre-trained models such as BART-base and T5-base. Another limitation stems from the exclusive focus of experiments on \datasetee{} and \datasetdart{}. To overcome these limitations and enhance the robustness of our model, it is crucial to broaden our evaluation scope. This involves extending assessments to larger models like BART-large and exploring the performance across a spectrum of large language models, including Alpaca~\cite{taori2023stanford}, ChatGPT~\cite{wu2023brief}, Flan-T5~\cite{longpre2023flan}, and Llama~\cite{touvron2023llama}. Furthermore, testing our model with various NLG datasets, such as WebNLG and Totto, is imperative for a more comprehensive understanding of its capabilities and limitations in various domains.

Although our self-training model diminishes the volume of training data required, we are curious about the training time of self-training methods versus full training methods. It remains uncertain whether the inference time of self-memory contributes to an extended training duration. This limitation is solved simply by measuring the training time of those methods.

Lastly, we currently lack a mechanism to regulate the ratio between self-mem and new data in self-training T2D and D2T models. Our current approach involves maintaining a consistent data distribution (30\%) in each training iteration. If self-memory data are deficient, we compensate by incorporating additional input data during inference to augment the dataset. However, this adjustment may result in prolonged training times.

\section{Conclusion}
We introduced a novel training model, STSM, designed for DTG problems. This model incorporates self-memory and newly acquired data to self-train D2T and T2D models. Through experiments conducted on two datasets, \datasetdart{} and \datasetee{}, our findings indicate that our model performs competitively compared to full data training when using less training data. We identified that the optimal method involves training on a mixture of self-memory and new data, incorporating self-training on the D2T model without needing self-training on the T2D model. It also shows that self-training on the T2D model is unimportant. Moreover, the question of whether training on fixed data is superior to training on random data remains uncertain.

Moving forward, we plan to conduct more in-depth investigations into the optimal rate combination of self-memory and new data for self-training. Additionally, we aim to test our model on other NLG datasets and large language models to assess self-memory effectiveness in the self-training process. Moreover, we are keen on exploring the intriguing approach of integrating self-memory with external data generated by ChatGPT.

\section*{Compliance with Ethical Standards}
\begin{itemize}
    \item This article does not contain any studies with human participants or animals performed by any of the authors.
    \item All authors certify that they have no affiliations with or involvement in any organization or entity with any financial interest or non-financial interest in the subject matter or materials discussed in this manuscript.
\end{itemize}

\section*{Data Availability Statement}
\newcommand*{\brokenurlwithoutpar}[2]{{\texttt{#1}}\\*{\texttt{#2}}}
The curated dataset is publicly available at:
\begin{itemize}
    \item \url{https://github.com/hoangthangta/STSM}
\end{itemize}

\section*{CRediT Author Statement}
\textbf{Hoang-Thang Ta}: Conceptualization, Methodology, Software, Validation, Investigation, Data curation, Writing – review \& editing, Visualization. 








\bibliographystyle{elsarticle-num-names}
\bibliography{elsarticle-template-num-names}


\end{document}